\documentclass{article}
\pdfoutput=1

\usepackage[nonatbib, preprint]{neurips_2023}

\usepackage[utf8]{inputenc} 
\usepackage[T1]{fontenc}    
\usepackage{hyperref}       
\usepackage{url}            
\usepackage{booktabs, multicol, multirow}       
\usepackage{amsfonts}       
\usepackage{nicefrac}       
\usepackage{microtype}      
\usepackage{xcolor}         

\usepackage[linesnumbered, ruled]{algorithm2e} 
\usepackage{algorithmicx}

\usepackage{enumitem}

\usepackage{amsmath}
\usepackage{graphicx}
\usepackage{capt-of}
\usepackage{multicol}
\usepackage{amsthm}
\usepackage{placeins}
\usepackage{array,xspace}

\title{LLMs Can Understand Encrypted Prompt: \\ Towards Privacy-Computing Friendly Transformers}

\author{%
  Xuanqi Liu \thanks{Work partially supported by an internship program funded by Sudo Privacy.} \\
  Tsinghua University \\ Sudo Privacy \\
  \texttt{lxq22@mails.tsinghua.edu.cn} \\
  \And
  Zhuotao Liu \thanks{Corresponding author.} \\
  Tsinghua University \\
  \texttt{zhuotaoliu@tsinghua.edu.cn} \\
}

\newcommand{\eg}{\emph{e.g.,}\xspace}
\newcommand{\ie}{\emph{i.e.,}\xspace}

\newcommand{\share}[1]{\langle #1 \rangle}
\newcommand{\tensor}[1]{\mathbf{#1}}
\renewcommand{\vector}[1]{\mathbf{#1}}
\newcommand{\ring}{\mathcal{R}}
\newcommand{\enc}[1]{[\![#1]\!]}
\newcommand{\field}{\mathbb{F}}
\newcommand{\ReLU}{\mathsf{ReLU}}
\newcommand{\GELU}{\mathsf{ReLU}}
\newcommand{\Softmax}{\mathsf{Softmax}}
\newcommand{\LayerNorm}{\mathsf{LayerNorm}}
\newcommand{\loss}{\mathcal{L}}

\newcommand{\protocol}{\Pi}
\newcommand{\protocolMatmul}{\protocol_{\mathsf{MatMul}}}
\newcommand{\protocolElemul}{\protocol_{\mathsf{EleMul}}}
\newcommand{\protocolMax}{\protocol_{\mathsf{max}}}
\newcommand{\protocolReLU}{\protocol_{\mathsf{ReLU}}}
\newcommand{\protocolExp}{\protocol_{\mathsf{exp}}}
\newcommand{\protocolRecip}{\protocol_{\mathsf{recip}}}
\newcommand{\protocolRSqrt}{\protocol_{\mathsf{rSqrt}}}
\newcommand{\protocolTanh}{\protocol_{\mathsf{tanh}}}

\newcommand{\crossterm}{\mathsf{cross}}

\newcommand{\model}{\mathcal{M}}
\newcommand{\modelAccept}{\model_{\mathsf{accept}}}
\newcommand{\modelTemp}{\model_{\mathsf{temp}}}
\newcommand{\evaluate}{\mathsf{Eval}}

\newcommand{\paraspace}{\vspace{0.01in}}
\newcommand{\parab}[1]{\paraspace\noindent{\bf #1.}}

\newcommand{\tabincell}[2]{\begin{tabular}{@{}#1@{}}#2\end{tabular}}

\newtheorem{definition}{Definition}

\begin{document}

\setlength{\belowdisplayskip}{0.3ex} \setlength{\belowdisplayshortskip}{0.3ex}
\setlength{\abovedisplayskip}{0.3ex} \setlength{\abovedisplayshortskip}{0.3ex}

\maketitle

\begin{abstract}
  The community explored to build private inference frameworks for transformer-based large language models (LLMs) in a server-client setting, where the server holds the model parameters and the client inputs its  private data (or prompt) for inference. However, these frameworks impose significant overhead when the private inputs are forward propagated through the original LLMs. In this paper, we show that substituting the computation- and communication-heavy operators in the transformer architecture with privacy-computing friendly approximations can greatly reduce the private inference costs while incurring very minor impact on model performance. Compared to state-of-the-art Iron (NeurIPS 2022), our privacy-computing friendly model inference pipeline achieves a $5\times$ acceleration in computation and an 80\% reduction in communication overhead, while retaining nearly identical accuracy.
\end{abstract}

\section{Introduction}

Large language models (LLMs) attracted significant attentions, driven by advances in artificial intelligence and the availability of large amounts of training data~\cite{Openai2023gpt4, Brown2020language}. LLMs are trained on massive datasets of text and code, and can be used for a variety of tasks, including generating text, translating languages, writing different kinds of creative content, and answering questions in an informative way.

Nowadays, LLMs are usually provided as online inference services. This, however, raises serious privacy concerns. On the one hand, the client's input (prompt, such as questions and requirements) must be submitted in plaintext to the service provider. In certain use cases, the prompt could contain sensitive information that the client would like to hide from the service provider. The growing concern for privacy, particularly in the age of Web3.0~\cite{liu2021make}, and the enactment of privacy protection laws such as the GDPR, necessitate that privacy be a top priority when offering online LLM services.
On the other hand, the LLMs hosted by the service provider are proprietary, so it is critical to ensure that an adversarial client cannot obtain the model parameters during inference. 

The private inference paradigm of neural networks has recently emerged as a solution to the aforementioned problem~\cite{Dowlin2016cryptonets, Juvekar2018gazelle, Liu2017minionn, Mishra2020delphi, Huang2022cheetah, Hao2022iron, Rathee2020cryptflow2}. In this paradigm, the client submits an encrypted version of its input and works collaboratively with the service provider to obtain an encrypted inference result that can only be recovered by the client itself. The service provider cannot obtain any private information about the input. However, the efficiency of the private inference, especially on large neural networks, is extremely limited by the extensive use of Homomorphic Encryption (HE) and Secure Multiparty Computation (MPC) primitives on various expensive neural network operators.

Two recent art~\cite{Huang2022cheetah} and \cite{Hao2022iron} demonstrate the possibility of private inference on popular neural networks (\eg convolutional networks and transformers) in computer vision and natural language processing, respectively. We observe that the time and communication cost of private inference on LLMs consisting of multiple transformer blocks is much higher than that of the traditional convolutional networks
For instance, even on a model as small as BERT-Tiny~\cite{bhargava2021generalization}, \cite{Hao2022iron} takes $\sim$50 seconds and 2GB of communication for a single inference, while Cheetah \cite{Huang2022cheetah} can scale to ResNet-32~\cite{He2016deep} with 15 seconds and 0.11 GB communication for one inference. This difference is because transformers use sophisticated nonlinear functions that are \emph{computationally-unfriendly} to the cryptographic primitives. For instance, convolutional networks use ReLU~\cite{Fukushima1975cognitron} and batch normalization~\cite{Ioffe2015batch}, while transformers prefer GELU~\cite{Hendrycks2016gaussian}, meanwhile extensively use softmax function and layer normalization techniques~\cite{Ba2016layer}\footnote{During inference, batch normalization does not compute data statistics, but layer normalization does.}. Through experiments, we show that these functions take up to more than 70\% of the total cost for private inference on transformers.

This paper explores to improve the efficiency of private inference on transformer-based models. To this end, we first build a private inference system that fully supports the private computations required for transformer-based LLMs. We then conduct extensive experiments to identify the critical performance bottleneck. Based on this, we design various substitutions for these bottlenecked components, and use fine-tuning to retain model performance after replacing these components. Taken together, we build an effective system to provide LLM inference service while fully protecting the privacy of the input data. We perform extensive evaluations and show that applying our privacy-computing friendly operators in LLMs can reduce $\sim 80\%$ of the overall private inference time, while retaining nearly identical model accuracy.

\subsection{Related Work}

Private inference of neural networks was first proposed by~\cite{Dowlin2016cryptonets}. It demonstrates the feasibility of fully using Homomorphic Encryption (HE) to achieve non-interactive private inference. However, due to the linearity restriction of HE, every nonlinear function such as ReLU and MaxPooling must be replaced by linear or polynomial approximation. Works after~\cite{Dowlin2016cryptonets} primarily sought to use Secure Multiparty Computation (MPC) to deal with the nonlinear functions, and exploit the single instruction multiple data (SIMD) property of HE to accelerate the inference~\cite{Juvekar2018gazelle, Mishra2020delphi, Rathee2020cryptflow2}. A recent art Cheetah~\cite{Huang2022cheetah} proposes a special encoding method to encode vectors and matrices into HE polynomials, which achieves state-of-the-art performance in computing matrix-vector multiplication and convolutions. 
Iron~\cite{Hao2022iron} realizes that matrix-matrix multiplication (rather than matrix-vector multiplication) dominates in transformer-based inference, and therefore improves the 
vanilla polynomial encoding by introducing a blocking method that prioritizes the batch dimension. Despite the optimization, some of the non-linear functions (\eg GELU, softmax and layer normalization layers) are fundamentally expensive in private inference. For instance, Iron~\cite{Hao2022iron} reports that running a single inference on BERT-Tiny~\cite{bhargava2021generalization} requires 50 seconds time and 2GB transmission. 
Two recent studies explore replacing these fundamentally expensive non-linear functions with operators that are more friendly in private inference. For instance, Chen et al. \cite{Chen2022thex} use ReLU to substitute all non-linearities in a transformer, and relying on HE for linear operations. However, their architecture requires the ReLU functions to be executed \emph{in plaintext by the client}, which may reveal the proprietary model owned by the server. Li et al. \cite{li2022mpcformer}, on the other hand, use quadratic polynomial approximations for GELU and softmax. Yet, they rely on Trusted Third Party (TTP) to produce correlated randomness for MPC. This is inappropriate in practice because designing and certifying a TTP is an open problem. 



\section{Preliminaries}

\subsection{Transformer Architecture}
Transformers dominate the model architecture in the area of natural language processing since its birth~\cite{vaswani2017attention}. The state-of-the-art large language models (LLMs) typically  consist of an embedding layer, a transformer encoder stack with $n$ identical encoder layers, and a downstream task sub-model (a classifier model for predicting labels or a generative model for predicting the next token)~\cite{devlin2018bert, liu2019roberta, Brown2020language, radford2019language, Openai2023gpt4}. In this paper, for simplicity we ignore the embedding layer (\ie both the server and the client could produce the embeddings with respect to some input sentence), and focus on private inference on the transformer encoder stack and the downstream sub-model.

One transformer encoder layer consists of two main parts: the multihead-attention and the feed-forward layer. A residual structure and layer normalization layer are inserted after both parts. Formally, for the input $x$ to go through one transformer encoder layer:
\begin{equation}\begin{aligned}
x_1 & = \mathsf{LayerNorm}_1(x + \mathsf{MultiheadAttention}(x)) \\
y & = \mathsf{LayerNorm}_2(x_1 + \mathsf{FeedForward}(x_1))
\end{aligned}\end{equation}
The multihead-attention consists of an input projection (a fully connected layer), a softmax function and an output projection (also an FC), while the feed-forward layer consists of two fully connected layers and an activation function between them (usually GELU).

Thus, a private inference system on transformer-based models should support forward propagation of fully connected layers, softmax function, GELU function, and layer normalization with private input.  

\subsection{Cryptography  Primitives}\label{subsection:cryptographic-primitives}

To realize a private inference system for LLMs, we mainly use two cryptographic primitives.  

\parab{Homomorphic Encryption}
Homomorphic encryption supports  computation (addition and multiplication) over ciphertexts. We use the BFV fully homomorphic encryption cryptosystem based on the RLWE problem with residual number system (RNS) optimization~\cite{Fan2012bfv, Bajard2017RNSBFV}. Specifically, the BFV scheme is constructed with a set of parameters $\{N, t, q\}$ such that the polynomial degree $N$ is a power of two, and $t$, $q$ represent plaintext and ciphertext modulus, respectively. We let $t$ be chosen as a power of two, $2^\ell$. The plaintext space is the polynomial ring $\ring_{t, N} = \mathbb{Z}_t[X]/(X^N+1)$ and the ciphertext space is $\ring_{q,N}^2$. Homomorphism is established on the integer polynomial ring $\ring_{t, N}$, supporting addition and multiplication of polynomials in the encrypted domain. We denote the homomorphically encrypted ciphertext of $x$ as $\enc{x}$.

\parab{Secure Multiparty Computation}
We utilize additive secret-sharing scheme upon the field $\field = \mathbb{Z}_t$ (integers modulo $t$) with $t=2^\ell$, 
where an integer $x \in \field$ is shared between a pair of client and server (\ie $x = \share{x}_0 + \share{x}_1$)~\cite{shamir1979share}. LLMs typically involves decimal numbers rather than integers. To adapt to the BFV scheme and integer-based secret sharing, we use a fixed-point representation of decimal numbers~\cite{Rathee2020cryptflow2, Huang2022cheetah}.
A decimal $\tilde{x} \in \mathbb{R}$ is represented as an integer $x = \mathsf{Encode}(\tilde{x}) = \lfloor \tilde{x} \cdot 2^f \rfloor \in \mathbb{Z}$, with a precision of $f$ bits. After every multiplication, the precision inflates to $2f$, and a truncation is required to keep the original precision. Since we use $\mathbb{Z}_t$ rather than $\mathbb{Z}$, we require all intermediate results in their decimal form $\tilde{x} \in \mathbb{R}$ not to exceed $\pm t/2^{2f}$, to prevent overflow. In the rest of the paper, unless stated otherwise, all scalars and elements of tensors are in $\field = \mathbb{Z}_t$.

\subsection{Threat Model}
We consider a semi-honest threat model including two parties: a server holding all the model weights, and a client holding the inference input data. The model architecture is public. The two parties adhere to the protocols but are curious about the private information held by the other party (\ie the model weights and inference inputs).

\section{Approach}

We first build a fully functional framework for private inference of transformers and LLMs based on transformers, including all building blocks such as FC layers, ReLU, GELU activation functions, etc. We run real-world models within the framework and measure the inference cost of each kind of operation to determine the bottleneck of the end-to-end inference pipeline. Then, we transform these computationally and communication heavy layers or functions into cryptography-friendly ones, and fine-tune the model to retain the model accuracy during substitution. Finally, we test the post-tuned models with our private inference framework to evaluate their inference performance. 

\subsection{Private Transformer Inference}
We use the secret-sharing form of all intermediate outputs throughout the private inference procedure to protect the privacy of both the inputs and the model weights. Concretely, we treat every neural network operator $y_i = f_i(x_i)$ as a 2-party computation protocol, which takes secret-shares $x_i = \share{x_i}_0 + \share{x_i}_1$ from the two parties as input ant returns the secret-shares of $y_i = \share{y_i}_0 + \share{y_i}_1$ to them. All operators in the transformers could be divided into two categories: (1) linear operators (\eg fully connected layers); (2) non-linear operators (\eg GELU, Softmax function). We do not consider the residual structure as a single operator as it is simply an addition of secret shares.

\subsubsection{Linear Operators}
The core linear protocol for private inference over linear layers is the matrix multiplication protocol. It is realized using homomorphic encryption, with the polynomial encoding primitive first proposed by \cite{Huang2022cheetah} and extended by \cite{Hao2022iron}. We start with a simple situation where one party holds $\tensor{A}$ and the other party holds $\tensor{B}$. The protocol $\protocolMatmul(\tensor{A}, \tensor{B})$ takes the inputs from the two parties and outputs the secret shares $\share{\tensor{C}}_0, \share{\tensor{C}}_1$ such that $\tensor{C} = \tensor{A} \tensor{B}$. We suppose the input matrices are small enough to be encoded into one plaintext polynomial, as larger matrices could be split into smaller blocks to adapt the protocol. We summarize this protocol $\protocolMatmul$ in Algorithm~\ref{alg:private-matmul}.

\begin{algorithm}[t]
    \setlength{\belowdisplayskip}{0.3ex} \setlength{\belowdisplayshortskip}{0.3ex}
    \setlength{\abovedisplayskip}{0.3ex} \setlength{\abovedisplayshortskip}{0.3ex}
    \caption{Private matrix multiplication protocol $\protocolMatmul$}
    \label{alg:private-matmul}
    \KwIn{The server inputs $\tensor{A} \in \mathbb{R}_{m\times r}$ and the client inputs $\tensor{B} \in \mathbb{R}_{r\times n}$, $mrn \leq N$, $N$ being the BFV polynomial degree.}
    \KwOut{The two parties receives the secret shares of $\tensor{C}=\tensor{A}\tensor{B}$.}
    Server and client respectively encode $\tensor{A}=(a_{ij}), \tensor{B}=(b_{jk})$ into polynomial $a = \pi_A(\tensor{A}), b = \pi_B(\tensor{B})$: 
    $$a = \textstyle\sum_{i=0}^{m-1}\textstyle\sum_{j=0}^{r-1} a_{ij}x^{ir+r-1-j}, b = \textstyle\sum_{j=0}^{r-1}\textstyle\sum_{k=0}^{n-1} b_{jk}x^{kmr + j}$$ \\
    Client encrypts the polynomial $b$ and sends $\enc{b}$ to server. \\
    Server use HE to evaluate $\enc{c}=a \cdot \enc{b}$, and samples a random polynomial $\share{c}_1 = s$. Server sends $\enc{\share{c}_0}=\enc{c}-s$ to client for decryption. \\
    Server and client respectively output $\share{\tensor{C}}_i=\pi_C^{-1}(\share{c}_i)$, where the decoding method $\pi_C^{-1}$ for $c=\sum_{i=0}^{N-1}c_ix^i$ is: (note that only a part of the coefficients in $c$ is used)\label{alg:private-matmul:step:decode-c}
    $$\tensor{C} = \pi_C^{-1}(c) = (c_{kmr + ir + r - 1})_{ik}.$$
\end{algorithm}


\parab{Fully connected layer} In fully connected layers ($\vector{y} = f(\vector{x}) = \tensor{W}\vector{x} + \vector{b}$), we can directly use $\protocolMatmul$ in the forward propagation:
\begin{itemize}[itemsep=0pt,leftmargin=2em,topsep=0pt]
    \item Suppose the client holds $\share{\vector{x}}_0$, while the server holds $\share{\vector{x}}_1$, the weights $\tensor{W}$ and bias $\vector{b}$.
    \item The two parties invoke $\protocolMatmul(\tensor{W}, \share{\vector{x}}_0)$ to produce $\share{\tensor{W}\share{\vector{x}}_0}_0$ and $\share{\tensor{W}\share{\vector{x}}_0}_1$.
    \item The client outputs $\share{\vector{y}}_0 = \share{\tensor{W}\share{\vector{x}}_0}_0$, and the server outputs $\share{\vector{y}}_1 = \share{\tensor{W}\share{\vector{x}}_0}_1 + \tensor{W}\share{\vector{x}}_1 + \vector{b}$.
\end{itemize}

\parab{Attention} In multihead attention, however, we need to calculate the multiplication of two secret-shared matrices (\ie $\tensor{Q}\tensor{K}^T$ and $\tensor{A}\tensor{V}$, for $\tensor{A}=\mathsf{Softmax}(\tensor{Q}\tensor{K}^T/\sqrt{E})$). The key observation is that we need only to calculate the secret-shares of the ``cross terms'', by invoking the $\protocolMatmul$ protocol twice. Suppose we need to compute $\tensor{Z} = \tensor{X}\tensor{Y}$, with both the input matrices secret-shared. The two parties perform the following procedure:
\begin{itemize}[itemsep=0pt,leftmargin=2em,topsep=0pt]
    \item The two parties invoke $\protocolMatmul(\share{\tensor{X}}_1, \share{\tensor{Y}}_0)$ and $\protocolMatmul(\share{\tensor{X}}_0, \share{\tensor{Y}}_1)$\footnote{For clarity, we semantically abuse the notation: to make sure that the client encrypts/decrypts and the server performs HE operations, the correct form should be $\protocolMatmul(\share{\tensor{Y}^T}_1, \share{\tensor{X}^T}_0)$, and the two parties transposes the result after the invocation.} and add their results, so that they obtain the secret shares of the cross terms:
    \begin{equation}\share{\tensor{Z}_{\crossterm}}_0 + \share{\tensor{Z}_{\crossterm}}_1 = \tensor{Z}_{\crossterm} = \share{\tensor{X}}_1 \share{\tensor{Y}}_0 + \share{\tensor{X}}_0 \share{\tensor{Y}}_1\end{equation}
    \item The client outputs $\share{\tensor{Z}}_0 = \share{\tensor{Z}_{\crossterm}}_0 + \share{\tensor{X}}_0 \share{\tensor{Y}}_0$; The server outputs $\share{\tensor{Z}}_1 = \share{\tensor{Z}_{\crossterm}}_1 + \share{\tensor{X}}_1 \share{\tensor{Y}}_1$.
\end{itemize}

\subsubsection{Non-linear Operators}
For the non-linear operators, we mainly use several primitives provided by~\cite{Huang2022cheetah, Rathee2020cryptflow2} libraries, which rely on the oblivious transfer cryptographic primitive. Recall that each operator takes as input secret-shares, and output secret-shares to the two parties. We use these primitives as black boxes in our system:

\begin{minipage}{\textwidth}
\begin{multicols}{2}
    \begin{itemize}[itemsep=0pt,leftmargin=2em,topsep=0pt]
        \item $\protocolReLU$: $\mathsf{ReLU}(x) = \max\{x, 0\}$.
        \item $\protocolElemul$: field element multiplication $x \cdot y$.
        \item $\protocolMax$: $\max(\vector{x}) = \max\{x_i | \vector{x} = (x_i)\}$.
        \item $\protocolExp$: $\mathsf{exp}(x) = e^x$, for $x \leq 0$.
        \item $\protocolRecip$: $\mathsf{recip}(x) = 1/x$.
        \item $\protocolRSqrt$: $\mathsf{rSqrt}(x) = 1/\sqrt{x}$, for $x > 0$.
        \item $\protocolTanh$: $\tanh(x) = \frac{1-e^{-2x}}{1+e^{-2x}}$.
    \end{itemize}
\end{multicols}
\end{minipage} \par

We now discuss the private inference procedure for each kind of non-linearity in the transformer architecture.

\parab{GELU} GELU is an activation function commonly used in transformers:
$$\mathsf{GELU}(x) = 0.5x\cdot\left[1 + \tanh\left(\sqrt{2/\pi}(x+0.044715x^3)\right)\right]$$
To produce $\mathsf{GELU}(x)$ for secret shared $\share{x}$, the two parties invoke $\protocolElemul(\share{x}, \share{x})$ to produce $\share{x^2}$, and again invoke $\protocolElemul(\share{x}, \share{x^2})$ to produce $\share{x^3}$. Addition and multiplication by scalar are performed subsequently before invoking $\protocolTanh$ on $\share{\sqrt{2/\pi}(x+0.044715x^3)}$. Finally, they again invoke once $\protocolElemul$ to obtain the final result $\share{\mathsf{GELU}(x)}$.

\parab{Softmax} Softmax is a key operator in scaled-dot attention construction. It is applied to the attention scores as a non-linearity and normalization to put more verbosity into the model. For a vector $\vector{x} = (x_0, \cdots, x_{n-1})$,
$$\mathsf{Softmax}(\vector{x}) = \left(e^{x_i} / \textstyle\sum_{j=0}^{n-1}e^{x_j}\right)_{i\in[0, n)}$$
To compute the softmax function, the two parties first invoke $\protocolMax(\share{\vector{x}})$ to obtain $\share{x_{\max}}$. They substract the original vector by the maximum value to ensure the inputs to $\protocolExp(\share{\vector{x}-x_{\max}})$ is negative. The exponentiated results are summed and used in $\protocolRecip$ to produce the denominator in softmax, and finally they call $\protocolElemul$ to obtain $\share{\mathsf{Softmax}(\vector{x})}$.

\parab{LayerNorm} Layer normalization is an operator used to limit the bound of the layer outputs of self attention and feed forward sub-networks. It first calculate the mean and variance (standard deviation) along the embedding dimension and normalizes the input with these statistics, and then perform a learnable affine projection (with parameters $\gamma, \beta$) to produce the output ($\epsilon$ is a small term to prevent division by zero): 
$$\mathsf{LayerNorm}(\tensor{x}) = \frac{\tensor{x} - \bar{\tensor{x}}}{\sqrt{\mathrm{Var}(\tensor{x}) + \epsilon}} \cdot \gamma + \beta$$
For calculating the normalized values, the two parties invoke the $\protocolRecip$ and $\protocolRSqrt$ protocols, and for the affine projection, they invoke $\protocolElemul$ to produce the outputs. 

\subsubsection{Other Optimizations}

\parab{Reducing the communication cost of matrix multiplication} 
We observe that in the BFV homomorphic encryption system, the core of decryption could be rendered as computing $m=s\cdot c_1 + c_0$, where $c_0,c_1,s$ are all polynomials, and $c=(c_0,c_1)\in \ring_{q,N}^2$ is the HE ciphertext, $s$ the secret key. This form of decryption indicates that, to obtain each coefficient $m_i$ in $m$ there is only one coefficient $(c_0)_i$ required in $b$ but all coefficients in $c_1$ are needed. In Step~\ref{alg:private-matmul:step:decode-c} of Algorithm~\ref{alg:private-matmul}, to decipher $\tensor{C}$, only part of the cofficients of $c$ are required. Therefore, we could omit transmitting the other coefficients in the $c_0$ part of the ciphertexts. Since the required coefficients are very sparse in matrix multiplication, this roughly reduces half of the server-to-client communication for linear operators.

\parab{Hardware parallelization of HE operations} In RLWE-based HE cryptosystem, the operations are essentially done on the polynomial ring. We observe that the addition and multiplication of polynomials could be effectively parallelized.\footnote{Multiplication of polynomials is done by number theory transform (NTT) and elementwise multiplication.} Therefore, we implement a GPU-version of the BFV cryptosystem, including homomorphic addition and cipher-plain multiplication, to further accelerate linear evaluations.

\subsection{Identifying Performance Bottleneck}

\begin{figure}[t]
    \begin{minipage}{0.35\textwidth}
        \centering
        \small
        \begin{tabular}{c|rr}
            \toprule
            Operator & Time & Comm. \\
            \midrule
            MatMul    &  3.75s &  111MB \\
            Softmax   &  5.95s &  518MB \\
            GELU      &  3.67s & 1020MB \\
            LayerNorm &  0.62s &  165MB \\
            Total     & 13.99s & 1814MB \\
            \bottomrule
        \end{tabular}
        \captionof{table}{Inference cost of various operators used in BERT-Tiny}
        \label{tab:performance-bottleneck}
    \end{minipage}\hfill
    \begin{minipage}{0.55\textwidth}
        \centering
        \includegraphics[width=1\linewidth]{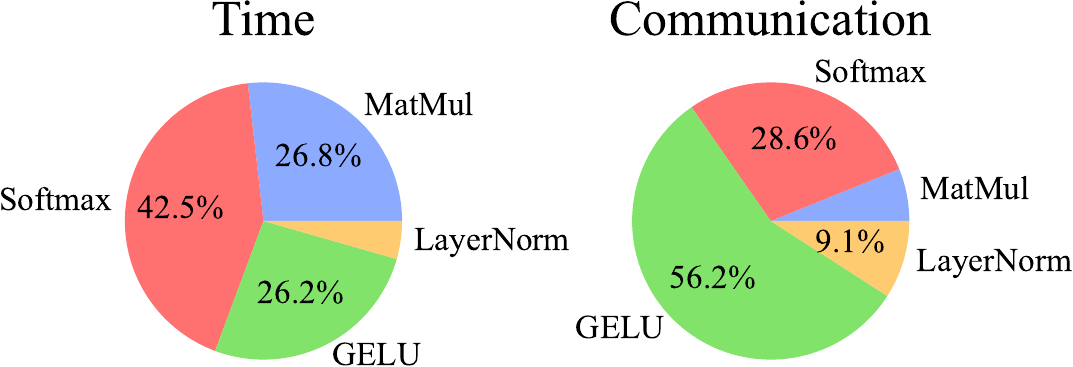}
        \caption{Ratio of various operators' cost}
        \label{fig:decomposition_cost}
    \end{minipage}
\end{figure}

Given the inference framework, we run private inference on the transformer-based language models to identify the performance bottleneck. As an example, we experiment with the BERT-Tiny~\cite{bhargava2021generalization, turc2019wellread} model with embedding dimension $E=128$, consisting of $n=2$ transformer encoder blocks. The maximum sequence length is set to 128, with shorter sentences padded. As a common practice, the embedding dimension of each attention head is $64$ (2 heads for $E=128$), and the hidden dimension of the feed forward block is $4E$. The time and communication costs are summarized in Table~\ref{tab:performance-bottleneck} and Figure~\ref{fig:decomposition_cost}.

These results indicate that the non-linear functions (GeLU, Softmax and LayerNorm) consume a significant portion of time and communication cost in the privacy inference pipeline, indicating that these non-linearities are \emph{not privacy-computing friendly}. For example, compared with ReLU activation function, GELU requires four evaluation of element-wise multiplication, and a computation-heavy $\tanh$ function based on look-up tables. 
Thus, it is critical to substitute these operators for privacy-computing friendly ones. Yet, retaining the model accuracy after applying alternative operators is non-trivial. 
In the following subsection, we elaborate on an automatic substitution workflow. 



\subsection{Exploring Privacy-computing Friendly Transformers}


In order to accelerate the private inference of transformers, the server substitutes the operators in its model with privacy-computing friendly alternatives and fine-tune the model to adapt to the replacement. Specifically, it replaces the GELU, Softmax and LayerNorm operators layer by layer, and test the model accuracy of each replacement after fine-tuning. The modification is accepted if the accuracy drop is within a predetermined threshold, or reverted otherwise.

\subsubsection{Substitution Workflow}\label{subsection:substitution-workflow}

We introduce a workflow to substitute the privacy-computing unfriendly operators in the transformer architecture. 
As introduced in preliminaries, the transformer-based language models typically include an encoder stack consisting of $n$ blocks of \emph{transformer encoder block} (in some literature called \emph{layers}), and the construction of each block is the same. Directly replacing all undesired operators with alternatives would greatly harm the model accuracy performance. 
Thus, we design a workflow to gradually substitute these operators layer by layer, from the last block to the first one. The model is fine-tuned between every substitution to make sure the model can adapt to the change. Denote the evaluate function for model $\model$ as $\evaluate(\model)$ (\ie testing the model on the validation data set) 
The workflow is summarized in Algorithm~\ref{alg:substitution-workflow}.

\begin{algorithm}[t]
    \caption{Substitution workflow}
    \label{alg:substitution-workflow}
    \DontPrintSemicolon
    \KwIn{An original transformer model $\model$ with $n$ encoder blocks $\{m_i\}_{i\in[n]}$; target original operator type $t$, and the replaced operator type $t'$; acceptable accuracy drop $\Delta \alpha$}
    \KwOut{The replaced model $\model'$}
    Evaluate the original model: $\alpha = \evaluate(\mathcal{M})$ and set $\modelAccept \leftarrow \model$\\
    \For{$i = n-1, n-2, \cdots 0$}{
        $\modelTemp \leftarrow$ substitute $t_i$ in the $i$-th block of $\modelAccept$ with $t'_i$. \\
        Finetune $\modelTemp$ with the parameters of block $0$ to $i-1$ fixed. \\
        Evaluate $\modelTemp$: $\alpha_i = \evaluate(\modelTemp)$. \\
        \lIf {$\alpha_i > \alpha - \Delta \alpha$}{
            $\modelAccept \leftarrow \modelTemp$.
        }
    }
    Output $\modelAccept$.
\end{algorithm}

\parab{Bound Controlling}
In the private inference framework, we adapt fixed-point decimals, where each real number is encoded into a integer with a scale of $2^f$. If the total bit-length is $\ell$, we have a plaintext space of $\ell-2f$ bits, because after each multiplication the scale will grow to $2^{2f}$. Our preliminary experiments show that when the encoder stack consists of many transformer encoder blocks, the absolute bound of the intermediate hidden states becomes larger after each block. As a result, the private inference procedure will encounter overflow in the secret-shares, producing meaningless prediction results. 

To address this issue, rather than directly using division to control the bounds, we modify the fine-tuning process to be aware of bound controlling. Specifically, we set an acceptable bound $B$, and add a loss term to penalize the hidden states with absolute values greater than $B$. Suppose for some sample $\vector{x}$, the hidden states of the transformer blocks are $\tensor{h}_i, i = 0, \cdots, n-1$. We design the loss function with three terms to be minimized during fine-tuning:
\begin{equation}
    \label{equation:loss-function}
    \loss = (1-\alpha_1-\alpha_2)\loss_{\mathrm{task}} + \alpha_1\loss_{\mathrm{decay}} + \alpha_2\loss_{\mathrm{bound}}
\end{equation}
where $\loss_{\mathrm{task}}$ is the loss (\eg cross entropy) function for the downstream task, $\loss_{\mathrm{decay}}$ is the weight decay term against overfitting, and 
$$\loss_{\mathrm{bound}} = \textstyle\sum_{i=0}^{n-1}||{\max\{|\tensor{h}_i| - B, 0\}}||_2$$
is a bounding term that penalizes the values too great. $|\tensor{h}_i|$ is taking the absolute value, and $||\cdot||_2$ is the L2-norm. Note that we do not directly set $\loss_{\mathrm{bound}}$ as the L2-norm of the hidden state, because we do not wish the values to be \emph{as small as possible}, but only require them to be \emph{lower than some bound} $B$. If the hidden states' values were too small, the relative error due to fixed-point approximation would become too large.

\subsubsection{Substitution Strategy}

As linear projection and ReLU activation function are more friendly to privacy computing, we mainly use these two components and their combination as our substitution candidates.

\parab{LayerNorm} The expensive part of the layer normalization operator is division operation of the standard deviation. Intuitively, the mean value is substracted to keep the intermediate activations centralized, and the deviation division is to keep them bounded. The average value may vary greatly across different samples, but the standard deviation (or the bound) can be captured by the affine transformation $\hat{\tensor{x}}\cdot \gamma + \beta$. Based on this insight, we remove the standard-deviation calculation part and only keep the centralization and affine transformation to be fine-tuned:
$$\mathsf{LayerNorm'}(\vector{x}) = (\vector{x} - \bar{\tensor{x}}) \cdot \gamma + \beta$$

\parab{Softmax} Replacing softmax function is more challenging. As the attention mask is added to the attention scores before the softmax, and the attention mask may contain $-\infty$ values, we cannot directly use a linear transformation ($\infty$ values would contaminate the intermediate results). Therefore, we combine the relatively cheap function ReLU to eliminate them. Furthermore, we notice that softmax function ensures that the output summed along the last dimmension is one. To simulate this feature, we divide the ReLU'ed result by their sum to obtain the output. Formally, for input $\vector{x}=(x_0,\cdots,x_{n-1})$ ($\epsilon$ is a small constant to prevent division by zero):
$$\mathsf{Softmax}'(\vector{x}) = ({\ReLU(x_i)} / {\textstyle\sum_{j=0}^{n-1}\ReLU(x_j)+\epsilon})_{i\in[0,n)}$$

Since softmax function itself is not trainable, whenever we substitute softmax function with the simplified version, we treat the input and output projections of the related multi-head attention as the trainable parameters in Algorithm~\ref{alg:substitution-workflow}. With this approach, the whole model can adapt to the softmax replacements faster.

\parab{GELU} GELU is the most expensive operator in the private inference pipeline. It is surprising that they could be simply replaced with ReLU, with nearly no accuracy drop in model performance.

\section{Evaluation}

We first evaluate the accuracies of our substituted models. Then we execute the entire private inference pipeline to measure the end-to-end privacy inference performance. 


\subsection{Experimental Setup}
\parab{Models and datasets} We test our substitution strategies and the private inference framework with three models: BERT-Tiny, BERT-Medium~\cite{bhargava2021generalization, turc2019wellread} and RoBERTa-Base~\cite{liu2019roberta} (referred to as Tiny, Medium and Base hereafter). These three models have similar architectures, only differing in hyperparameters (number of transformer encoder blocks $n=2,8,12$ and embedding dimensions $E=128,512,768$ respectively). During training, we limit the sequence length to $512$ tokens, while in inference we use a shorter $128$ length for efficiency. We use the MPRC, SST-2 and QNLI subsets of the GLUE benchmark~\cite{wang2018glue} to evaluate the accuracy performance of fine-tuned models.

\parab{Implementation} For plaintext fine-tuning, we implemented the substitution and training procedure with the Huggingface's \texttt{transformers} and the widely-applied \texttt{pytorch} libraries. For private inference, our 2-party interactive framework is build upon HE and MPC primitives. For the HE part we implement a GPU version of the BFV scheme~\cite{Fan2012bfv, Bajard2017RNSBFV}. We set polynomial degree $N=8192$, with $q$ set to $\sim 180$ bits. For the MPC primitives, we directly use the funtionalities provided by two open-sourced library: OpenCheetah~\cite{Huang2022cheetah} and SCI~\cite{Rathee2020cryptflow2}. The plaintext and secret-sharing modulus $t=2^{41}$, and bit precision $f=13$. We combine the HE and MPC cryptographic primitives using a high-level framework written in \texttt{python}, where we build different types of neural network layers as independent modules and provide end-to-end private inference interfaces. We evaluate our framework on a physical machine with Intel Xeon Gold 6230R CPU and NVIDIA RTX A6000 GPU (CUDA version 11.7).

\subsection{Operator Substitution}

We suspect that the more layers we substitute, the worse the model accuracy. Therefore, to largely retain the model performance, 
we substitute the layers starting from the most expensive ones to the least expensive ones. Specifically, denote the two layer normalizations in each block as LN1 and LN2, we first substitute all GELUs, then softmaxs, then LN1s and finally LN2s. 

To prevent a too strict bound from impacting the fine-tuning accuracy, we introduce gradually decreasing controlling bound (Section~\ref{subsection:substitution-workflow}) in these four substitutions: we set the acceptable bound $B=+\infty$ (\ie no bound controlling) when replacing GELUs; $B=32$ replacing softmaxes; $B=24$ replacing LN1s, and finally $B=16$ replacing LN2s. This setting allows the model to gradually adapt to smaller values of intermediate hidden states in the fine-tuning procedure. We set the maximum acceptable accuracy drop to $\Delta \alpha = 2\%$. In the loss function (Eq.~\ref{equation:loss-function}), we set $\alpha_1 = 0.1$ and $\alpha_2=0.2$.

\begin{figure}[t]
    \centering
    \includegraphics[width=0.9\textwidth]{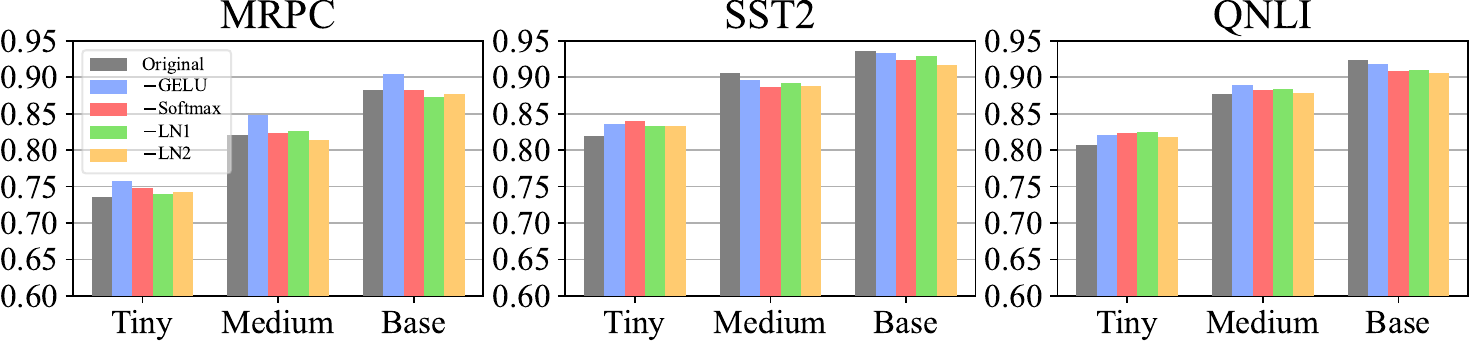}
    \caption{Accuracies of the model before and after each substitution. We use $-$ mark to denote which operators have been replaced. The changes are done incrementally. For example, ``$-$Softmax'' means both GELU and Softmax are replaced with privacy-computing friendly alternatives.}
    \label{fig:accuracies}
\end{figure}

After each substitution, we test the accuracy of the model. The results are shown in Figure~\ref{fig:accuracies}.
Overall, we observe that our substitution strategies can successfully replace all the GELU and softmax functions in every model, with accuracy drop of $\Delta \alpha < 2\%$. But only part of the LN layers in larger models can be replaced without significant accuracy drop. We notice that earlier LN layers are more important. 
For example, in the MRPC task, substituting the later 10 LN2s of BERT-Medium only accuracy by  $<1\%$. Yet once the first 2 layers of BERT-Medium is changed, the model cannot converge .
Interestingly, we observe that using ReLU instead of GELU sometimes results in better accuracy than the original models. 
This might be because that GELU is more helpful when training (possibly unsupervised) from scratch due to its non-zero differentials in the negative domain, yet the complexity of GELU may not be  necessary when fine-tuning for a downstream task. 


\subsection{End-to-end Performance}

\begin{figure}[t]
    \begin{minipage}{0.72\textwidth}
        \centering
        \scriptsize
        \begin{tabular}{c|c|r|rrrrr|r}
            \toprule
            & & Iron & \multicolumn{5}{r|}{Ours} & \multirow{2}{*}{\tabincell{r}{Impro-\\vement}} \\
            Model & Cost & \cite{Hao2022iron} & Orig. & $-$GE. & $-$Sm. & $-$LN1 & $-$LN2  \\
            \midrule
            BERT
            & Time       & 26.24 & 13.89  & 8.75  & 3.35  & 3.10  & 2.86 & 9.2$\times$ \\
            -Tiny
            & Comm.      &  1.07 &  1.77  & 0.78  & 0.33  & 0.27  & 0.22 & 4.9$\times$ \\
            \midrule
            BERT
            & Time      & 108.53 & 60.99  & 42.16 & 19.84 & 19.35 & 19.08 & 5.7$\times$ \\
            -Medium
            & Comm.      &   4.23 &  7.03  &  3.06 &  1.25 &  1.05 &  0.85 & 5.0$\times$ \\
            \midrule
            RoBERTa
            & Time       & 168.43 & 84.50  & 59.19 & 35.54 & 34.72 & 35.01 & 4.8$\times$ \\
            -Base
            & Comm.      &   6.38 & 9.51  &  3.55 &  1.74 &  1.44 &  1.14  & 5.6$\times$ \\
            \bottomrule
        \end{tabular}
        \captionof{table}{Private inference costs on our privacy-computing friendly models. Time costs are in seconds, and communication costs are in GB. ``Orig.'', ``GE.'' and ``Sm.'' stands for Original, GELU and Softmax.}
        \label{tab:costs}
    \end{minipage}\hfill
    \begin{minipage}{0.25\textwidth}
        \centering
        \includegraphics[width=1\linewidth]{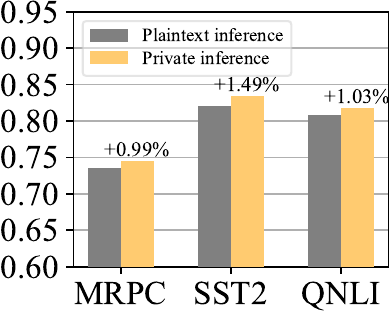}
        \caption{End-to-end model accuracy of private inference.}
        \label{fig:plain-enc-comparison}
    \end{minipage}
\end{figure}

We measured the runtime and communication cost of the end-to-end private inference on our privacy-computing friendly models. 
We first report the cost of \emph{one single} encoder layer. The results are shown in Table~\ref{tab:costs}. Since the GELU and softmax are the most expensive operators in the privacy inference, replacing these two operators with alternative operators results in $3\times$ speedup and reduction of 80\% communication cost. Furthermore, when the LN layers are replaced with affine transformations, the communication costs are further reduced to 13\% of the original model. 
We also compared the efficiency of our implementation to Iron~\cite{Hao2022iron}, where our approach showed state-of-the-art runtime and communication cost in private inference of LLMs, outperforming Iron by five times in runtime and communication efficiency. Note that when using the original model, our communication costs are slightly greater than Iron's because we use larger parameters in HE and MPC to retain high precision in large models.

Further, we report the end-to-end accuracy of private inference on BERT-Tiny for the three datasets. As shown in Figure~\ref{fig:plain-enc-comparison}, the private inference on our private-computing friendly model achieves slightly better accuracies compared to plaintext inference on the original model. This result is consistent with Figure~\ref{fig:accuracies}, where our modified BERT-Tiny performs slightly better than the original model on plaintext inference. 


\section{Discussion}


\parab{Model Pruning and Knowledge Distillation} The overhead of private inference is still much higher than plaintext inference, and the overhead is proportional to the model size (number of transformer encoder blocks and embedding dimension). Thus, 
it might be beneficial to consider model pruning and knowledge distillation~\cite{Michel2019aresixteen, Lagunas2021block, sun2019patient} to reduce the scale of the model while retaining comparable model accuracy.

\parab{Hardware Assist in MPC} Hardware acceleration is widely adopted in the field of machine learning. 
In this work, we have explored the feasibility of using parallel hardware to accelerate HE operations. 
Prior works proposed to utilize similar parallelization techniques to accelerate MPC primitives ~\cite{Knott2021crypten, Tan2021cryptgpu}. Incorporating these techniques might further reduce the inference costs.

\parab{Trusted Hardware} 
To further accelerate cryptography operations, it is possible to rely on trusted hardware.  
Prior works proposed to generate related randomness (\eg distributing beaver triples, random oblivious transfers)~\cite{Knott2021crypten, li2022mpcformer} using application-specific trusted hardware or generic Trusted Execution Environment (\eg Intel SGX)~\cite{Zhou2022ppmlac,Zheng2020asurvey}. These trusted hardware essentially serves as a Trusted Third Party (TTP). In future work, we will explore how TTP may further accelerate private inference in the case where the introduction of a TTP is acceptable by all stakeholders. 

\parab{Malicious Security} We currently build our private inference framework with passively secure protocols that protect privacy only from semi-honest adversaries. To strengthen our framework to malicious security, one could apply extra consistency checking protocols (\eg \cite{Keller2015actively, Yang2020Ferret} uses extra checks to ensure the fundamental oblivious transfers are generated correctly). 
Additionally, the advances in zero-knowledge proof (ZKP) technology, especially the development of zero-knowledge succinct non-interactive arguments of knowledge (zk-SNARK), sparked significant research on how to force desired participant behavior by requiring the participants to prove their behavior in a zero-knowledge and efficient manner using zero-knowledge proving systems. For instance, \cite{Cramer2022SPDZ2kEM} and~\cite{Li2023Efficient} apply ZKP over extension fields and large prime fields respectively to guarantee the faithful execution of the multiplication protocol. 
martFL~\cite{martfl} ensures that the training process of Federated Learning is fair to data trading. 

\section{Conclusion}

We propose an efficient framework for private inference on large language models (LLMs). Homomorphic encryption (HE) and secure multi-party computation (MPC) are used respectively for linear and non-linear operators. We observe that the privacy-computing unfriendly operators are the performance bottleneck, and substituting them with privacy-computing friendly alternatives brings 5x acceleration and 80\% reduction of communication costs while retaining model accuracies. We hope this work will shed light on a practical way of adapting LLMs to offer privacy-preserving inference service.

\bibliographystyle{plain}
\bibliography{main}

\section{Supplementary Material}

\newcommand{\simulator}{\mathsf{Sim}}
\newcommand{\view}{\mathsf{View}^{\Pi}}
\newcommand{\compindist}{\approx_{\mathrm{c}}}
\newcommand{\variance}{\mathrm{Var}}

\subsection{Formal Description of Threat Model}

We provide a formal description of the threat model of two semi-honest parties in the private inference framework.
\begin{definition}
    A protocol $\Pi$ between a server holding model weights $\tensor{W}$ (the model architecture $\mathcal{M}$ is public) and a client holding inference data $\vector{x}$ is a \textbf{private inference protocol} if it satisfies the following guarantees.
    \begin{itemize}[itemsep=0pt,leftmargin=2em,topsep=0pt]
        \item \textbf{Correctness.} On every set of model weights $\tensor{W}$ and every input data $\vector{x}$, the output of the client is a prediction result $\vector{y}$ produced by correctly doing neural network inference on $\vector{x}$, \ie $\vector{y}=\mathcal{M}(\vector{x}; \tensor{W})$, and the output of server is $\perp$.
        \item \textbf{Security.} \begin{itemize}[itemsep=0pt,leftmargin=2em,topsep=0pt]
            \item (\textbf{Data privacy}) A corrupted, semi-honest server does not learn anything useful about the client's inference data $\vector{x}$. Formally, we require the existence of an efficient simulator $\simulator_S$ such that $\view_S \compindist \simulator_S(\tensor{W}, \perp)$, where $\view_S$ denotes the view of server in the execution of $\Pi$.
            \item (\textbf{Model privacy}) A corrupted, semi-honest client does not learn anything useful about the server's model weights $\tensor{W}$. Formally, we require the existence of an efficient simulator $\simulator_C$ such that $\view_C \compindist \simulator_C(\vector{x}, \vector{y})$, where $\view_C$ denotes the view of client in the execution of $\Pi$ and $\vector{y}$ denotes the output (inference result) of the protocol $\Pi$ to the client.
        \end{itemize}
    \end{itemize}
\end{definition}
The security proof of our private inference framework follows by the combination of each sub-protocol and the sequential composibility of each operators to a full transformer architecture. The security proof of matrix multiplication follows from the security of the RLWE-based BFV HE scheme~\cite{Fan2012bfv} and the proofs in~\cite{Huang2022cheetah, Hao2022iron} for the protocol itself. For the security proof of each non-linearity protocol, we refer the readers to~\cite{Rathee2020cryptflow2} for $\protocolElemul, \protocolExp, \protocolTanh$ and ~\cite{Huang2022cheetah} for $\protocolReLU, \protocolRSqrt, \protocolMax, \protocolRecip$. These protocols mainly relies on the security of oblivious transfer~\cite{Naor2001EfficientOT, Ishai2003Extending} and subfield vector oblivious evaluation~\cite{Yang2020Ferret} as basic cryptographic primitives.

\subsection{Detail of the Non-linear Protocols}
We provide detailed description of the three main non-linear protocols, GELU (Algorithm~\ref{alg:private-gelu}), Softmax (Algorithm~\ref{alg:private-softmax}) and LayerNorm\footnote{For simplicity, we assume the tensor is 2-dimensional. In transformer, usually the input is 3-dimensional with batch size, sequence and embedding dimensions. This could be coerced as 2-dimensional by squeezing all the dimensions except the last.} (Algorith~\ref{alg:private-layernorm}), in the private inference framework.

\begin{algorithm}[t]
    \caption{GELU evaluation on secret shares}
    \label{alg:private-gelu}
    \KwIn{The client inputs $\share{x}_0$ and the server inputs $\share{x}_1$, where $x=\share{x}_0 + \share{x}_1$.}
    \KwOut{The two parties receives the secret shares of $y = \GELU(x)$.}
    The two parties invoke $\protocolElemul(\share{x}, \share{x})$, followed by a truncation (see Section~\ref{subsection:cryptographic-primitives}), to produce shares $\share{x^2}$. \\
    The two parties invoke $\protocolElemul(\share{x}, \share{x^2})$, followed by a truncation to produce shares $\share{x^3}$. \\
    The two parties multiply shares $\share{x^3}$ locally with $\lfloor 0.044715\cdot 2^f\rceil$, and invoke a truncation to obtain $\share{0.044715x^3}$. \\
    The two parties add locally the shares $\share{x}$ and $\share{0.044715x^3}$, and multiply the sum with $\lfloor \sqrt{2/\pi} \cdot 2^f\rceil$, and truncates the product to obtain $\share{x'} = \share{\sqrt{2/\pi}(x + 0.044715x^3)}$. \\
    The two parties invoke $\protocolTanh$ to obtain $\share{\tanh(x')}$. \\
    The server adds $2^f$ to its share $\share{\tanh(x')}_1$. This step semantically produces the shares of $\share{1 + \tanh(x')}$. \\
    The two parties invoke $\protocolElemul(\share{x}, \share{1 + \tanh(x')})$, and truncates by $f+1$ bits (because of the $0.5$ term of $\GELU$), producing $\share{\GELU(x)}$.
\end{algorithm}

\begin{algorithm}[t]
    \caption{Softmax evaluation on secret shares}
    \label{alg:private-softmax}
    \KwIn{The client inputs $\share{\vector{x}}_0$ and the server inputs $\share{\vector{x}}_1$, where $\vector{x}=\share{\vector{x}}_0 + \share{\vector{x}}_1$.}
    \KwOut{The two parties receives the secret shares of $\vector{y} = \Softmax(\vector{x})$.}
    The two parties invoke $\protocolMax(\share{\vector{x}})$, obtaining $\share{x_\mathrm{max}}$. \\
    The two parties substract every element of $\share{\vector{x}}$ with $\share{x_\mathrm{max}}$ locally, and invoke $\protocolExp(\share{\vector{x}-x_\mathrm{max}})$, obtaining $\share{\vector{x}'} = \share{\exp(\vector{x} - x_\mathrm{max})}$. Denote the elements of $\vector{x}'$ as $(x'_i)_i$. \\
    The two parties take the sum along the secret-shared vector $\share{\vector{x}'}$ locally, producing shares $\share{s} = \share{\textstyle\sum_i x'_i}$. They invoke $\protocolRecip(\share{s})$ to produce $\share{1/s}$. \\
    The two parties invoke $\protocolElemul(\share{x'}_i, \share{1/s})$ for each element in $\vector{x}'$, followed by a truncation. This step produces the elements of $\share{\vector{y}} = \share{\Softmax(\vector{x})}$.
\end{algorithm}

\begin{algorithm}[t]
    \caption{Layer normalization evaluation on secret shares}
    \label{alg:private-layernorm}
    \KwIn{The client inputs 2d-tensor share $\share{\tensor{x}}_0$ and the server inputs $\share{\tensor{x}}_1$, where $\tensor{x}=\share{\tensor{x}}_0 + \share{\tensor{x}}_1$ is of shape $(N, E)$.}
    \KwOut{The two parties receives the secret shares of $\tensor{y} = \LayerNorm(\tensor{x})$.}
    The two parties locally takes the sum along the $N$ dimension, multiply the sum by $\lfloor 2^f/E \rceil$ and truncates it, obtaining the mean shares $\share{\bar{\vector{x}}}$.\label{alg-step:private-layernorm:step-mean} \\
    The two parties invoke $\protocolElemul(\share{\tensor{x} - \bar{\vector{x}}}, \share{\tensor{x} - \bar{\vector{x}}})$, followed by a truncation. Then they repeat a similar process as in~Step~\ref{alg-step:private-layernorm:step-mean} to produce the variance shares $\share{\variance(\tensor{x})}$. \\
    The server adds $\epsilon\cdot 2^f$ to his share $\share{\variance(\tensor{x})}_1$. This step semantically produces the shares of $\share{\variance(\tensor{x}) + \epsilon}$. \\
    The two parties invoke $\protocolRSqrt(\share{\variance(\tensor{X}) + \epsilon})$ to produce $\share{\vector{v}} = \share{1/\sqrt{\variance(\tensor{x}) + \epsilon}}$, and then they invoke $\protocolElemul(\share{\tensor{x} - \bar{\vector{x}}}, \share{\vector{v}})$ and a truncation, where $\share{\vector{v}}$ is repeated to fit the original shape $(N, E)$. This step produces the normalized values $\tilde{\vector{x}}$ shares, where $$\tilde{\vector{x}} = \frac{\tensor{x} - \bar{\vector{x}}}{\sqrt{\variance(\tensor{x}) + \epsilon}}.$$ \\
    For the affine transform, the two parties invoke $\protocolElemul(\share{\tilde{\vector{x}}}, \share{\gamma})$, where server provides $\share{\gamma}_1=\gamma$, and client provides $\share{\gamma}_0=0$. The two parties then invoke a truncation, and the server adds $\beta \cdot 2^f$ to its share. This step produces the final result $\share{\tensor{y}} = \share{\LayerNorm(\tensor{x})}$.
\end{algorithm}

\subsection{Additional Evaluation Details}

\parab{Codes} We make our codes publicly available at \url{https://github.com/privateLLM001/Private-LLM-Inference}.

\parab{Dataset details} We list details of the datasets used in our evaluation in Table~\ref{tab:dataset-details}.

\begin{table}[t]
    \centering
    \small
    \begin{tabular}{c|cc|rr}
        \toprule
        Name & Task & Domain & \#Train & \#Test \\
        \midrule
        MRPC & {2-class paraphrase} & News & 3.7k & 408 \\
        SST-2 & {2-class sentiment} & Movie reviews & 67k & 872 \\
        QNLI & {2-class question answering} & Wikipedia & 105k & 2k \\
        \bottomrule
    \end{tabular}
    \caption{Dataset details}
    \label{tab:dataset-details}
\end{table}

\parab{Accepted replacements} We list in Table~\ref{tab:model-replacements} the how many operators are successfully replaced in the evaluation for the three models and three datasets with our substution strategy, with the allowed accuracy drop set to $\Delta \alpha = 2\%$. We observe that the replacement for all GELUs and Softmaxes are successful, while in large models, a very small portion of the LN2s of the first few blocks could not be replaced. We conjecture that the first few layers are essential to capture the overall features of the input sentences, and thus play a vital role for high accuracies. We leave the exploration further into this phenomenon as a future work.

\begin{table}[ht]
    \centering
    \small
    \begin{tabular}{ccc|rrrr}
        \toprule
        Model & \#Blocks & Task & GELU & Softmax & LN1 & LN2 \\
        \midrule \multirow{3}{*}{BERT-Tiny} & \multirow{3}{*}{2}
          & MRPC & 2 & 2 & 2 & 2 \\
        & & SST2 & 2 & 2 & 2 & 2 \\
        & & QNLI & 2 & 2 & 2 & 2 \\
        \midrule \multirow{3}{*}{BERT-Medium} & \multirow{3}{*}{8}
          & MRPC & 8 & 8 & 8 & 8 \\
        & & SST2 & 8 & 8 & 8 & 8 \\
        & & QNLI & 8 & 8 & 8 & 8 \\
        \midrule \multirow{3}{*}{RoBERTa-Base} & \multirow{3}{*}{12}
          & MRPC & 12 & 12 & 12 & 10 \\
        & & SST2 & 12 & 12 & 12 & 12 \\
        & & QNLI & 12 & 12 & 12 & 11 \\
        \bottomrule
    \end{tabular}
    \caption{Successful model operator replacements for the three models and three datasets}
    \label{tab:model-replacements}
\end{table}

\end{document}